\documentclass{article}

\PassOptionsToPackage{numbers}{natbib}

\usepackage[preprint]{vv2}

\usepackage[utf8]{inputenc} 
\usepackage[T1]{fontenc}    
\usepackage{hyperref}       
\usepackage{booktabs}       
\usepackage{multirow}       
\usepackage{amsfonts}       
\usepackage{nicefrac}       
\usepackage{microtype}      
\usepackage{xcolor}         
\usepackage{graphicx}
\usepackage{listings}
\newcommand{\equalcontrib}{\footnotemark[1]}
\newcommand{\corresponding}{\thanks{Corresponding author}}

\newcommand{\widecenter}[1]{\makebox[\linewidth][c]{#1}}

\title{V{\Large ARCO}-V{\Large ISION}-2.0 Technical Report}
\newcommand\varcovision{{\normalsize V\small ARCO\normalsize-\normalsize V\small ISION}}

\author{%
  \begin{tabular}{@{}c c c@{}}
  Young-rok Cha\thanks{Equal contribution} & Jeongho Ju\equalcontrib & SunYoung Park \\
  Jong-Hyeon Lee & Younghyun Yu & Youngjune Kim\corresponding \\
  \end{tabular}\\[3pt]
  NC AI \\
  \texttt{\{jaycha,jeongho,sun0park,leejh1230,zrohyun\}@ncsoft.com} \\
  \texttt{datajuny@gmail.com} \\
}

\begin{document}

\maketitle

\begin{abstract}
We introduce \varcovision-2.0, an open-weight bilingual vision-language model (VLM) for Korean and English with improved capabilities compared to the previous model \varcovision-14B. The model supports multi-image understanding for complex inputs such as documents, charts, and tables, and delivers layout-aware OCR by predicting both textual content and its spatial location. Trained with a four-stage curriculum with memory-efficient techniques, the model achieves enhanced multimodal alignment, while preserving core language abilities and improving safety via preference optimization. Extensive benchmark evaluations demonstrate strong spatial grounding and competitive results for both languages, with the 14B model achieving 8th place on the OpenCompass VLM leaderboard\footnote{\url{https://rank.opencompass.org.cn/leaderboard-multimodal}} among models of comparable scale. Alongside the 14B-scale model, we release a 1.7B version optimized for on-device deployment. We believe these models advance the development of bilingual VLMs and their practical applications. Two variants of \varcovision-2.0 are available at Hugging Face: a full-scale 14B model\footnote{\url{https://huggingface.co/NCSOFT/VARCO-VISION-2.0-14B}} and a lightweight 1.7B model\footnote{\url{https://huggingface.co/NCSOFT/VARCO-VISION-2.0-1.7B}}.
\end{abstract}

\section{Introduction}

In December 2024, we released our first vision-language model (VLM), \varcovision-14B~\citep{varcovision}. While it achieved strong performance on many benchmarks compared to models of a similar size, it showed limitations in handling multi-image scenarios and Korean-localized tasks. To address these challenges, we present \varcovision-2.0, a new Korean-English VLM designed to understand both images and text and respond to user prompts with greater accuracy and fidelity.

With support for multi-image inputs, the model can effectively process complex visual content, such as documents, tables, and charts. It exhibits robust comprehension in both Korean and English, with notable advancements in Korean language generation and cultural contextual understanding. It shows improved performance across benchmarks and greater usability in practical scenarios, including general Q\&A, document parsing, and summarization. As of August 4, 2025, the 14B model ranks 8th on the OpenCompass VLM leaderboard among models with fewer than 20B parameters.

The training strategy follows a four-stage curriculum with memory-efficient techniques, yielding competitive results compared to other open-weight state-of-the-art (SoTA) models. \varcovision-2.0 demonstrates strong capabilities in spatial grounding and real-world perception, delivering high-quality OCR with text localization and robust performance on text-only tasks. To enhance accessibility, we additionally release a compact 1.7B variant, optimized for deployment on personal devices such as smartphones and PCs.

Major advancements include:
\begin{enumerate}
    \item \textbf{Multi-image Understanding:} Support for multi-image inputs allows the model to analyze multiple images simultaneously and make more holistic and context-aware decisions.
    \item \textbf{Korean Language Specialization:} The model is further specialized for Korean, with deeper understanding of Korean language, context, and culture. Korean text generation has been significantly improved, resulting in more natural, fluent, and accurate responses.
    \item \textbf{OCR with Text Localization:} \varcovision-2.0 can identify the position of the text and provide bounding boxes around it. This makes it especially useful for document understanding, signage interpretation, and structured visual data.
    \item \textbf{Enhanced Safety:} The model now offers improved handling of harmful or sexually explicit content, ensuring safer and more reliable interactions.
    \item \textbf{Two Model Types of \varcovision-2.0:} We release both a 14B full-scale model and a 1.7B lightweight variant of \varcovision-2.0 on Hugging Face, advancing bilingual VLM research and enabling practical on-device applications.
\end{enumerate}

Together, these features position \varcovision-2.0 as a practical, culturally adapted, open-access vision-language solution.

\begin{figure}[t]
  \label{fig:perf}
  \centering
  \includegraphics[width=\linewidth]{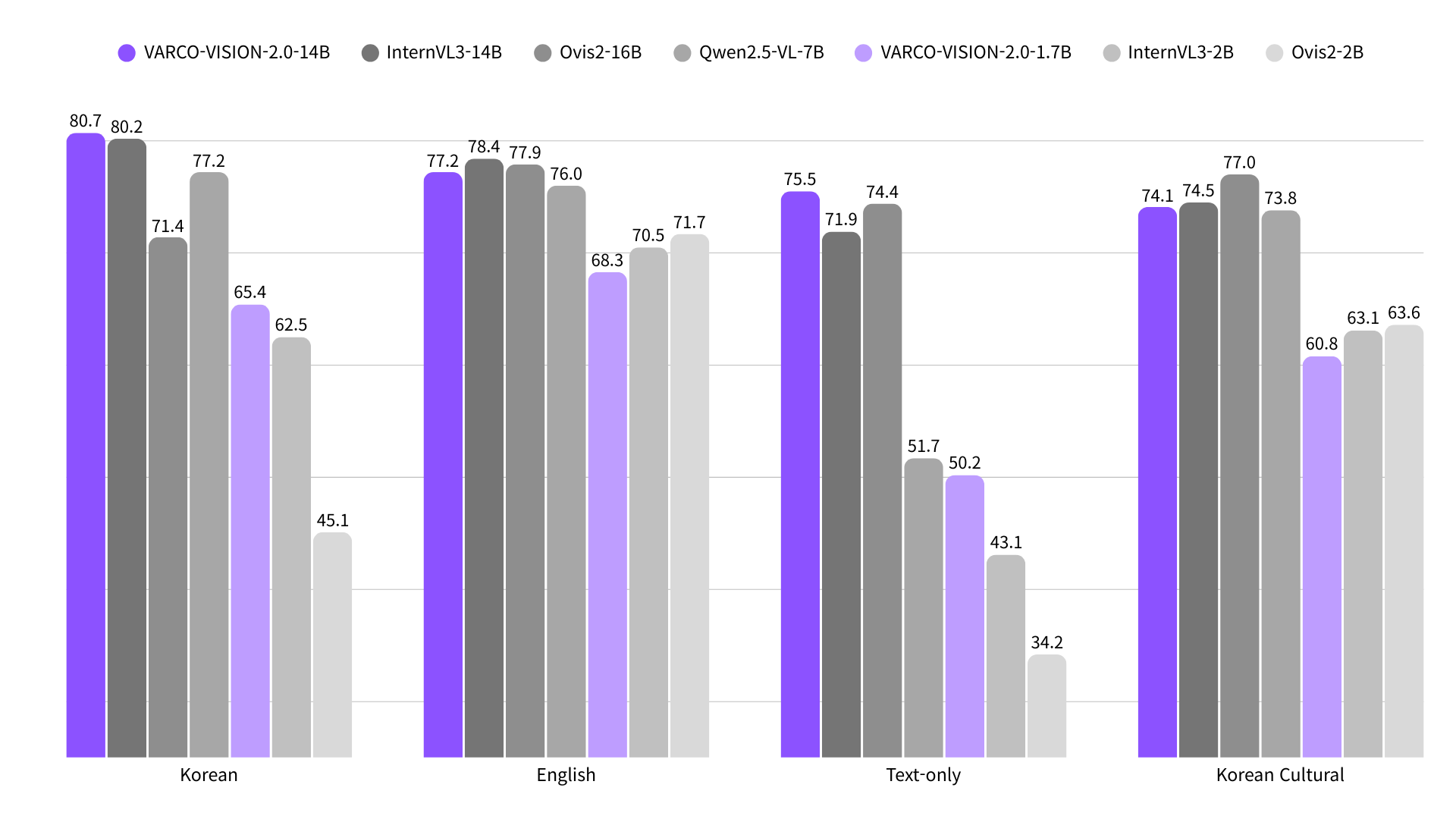}
  \caption{Average Performance of Models across Benchmark Categories.}
\end{figure}

\section{Training}

\subsection{Architecture}
\label{sec:arch}

\varcovision-2.0 is built on the LLaVA-OneVision~\citep{llavaonevision} architecture, combining a large language model (LLM), a vision encoder, and a two-layer MLP connector that projects image features into the LLM’s embedding space. We implement the model following the \texttt{Hugging Face Transformers} standard, which ensures immediate compatibility with the \texttt{transformers} ecosystem and enables deployment via \texttt{vLLM}~\citep{vllm} for production-scale inference without requiring additional code adaptation.

We adopt Qwen3~\citep{qwen3} as the LLM and SigLIP2~\citep{siglip2} as the vision encoder. For \varcovision-2.0, we employ SigLIP2 with a patch-16 configuration, replacing the SigLIP patch-14 setup used in \varcovision-14B. Prior work~\citep{llavaonevision} showed that token count has only a limited effect on overall quality of VLMs, motivating our use of the patch-16 setting. The patch-16 setting reduces the number of visual tokens compared to patch-14; for example, a 384$\times$384 input produces $24^2 = 576$ tokens with patch-16, considerably fewer than the $27^2 = 729$ tokens produced by patch-14. To process high-resolution images effectively, we apply the AnyRes strategy~\citep{llavaonevision}, which permits arbitrary input resolutions via tiled/cropped views with resolution-aware aggregation.

\subsection{Training Strategies and Datasets}
Our training pipeline follows a four-stage curriculum~\citep{qwen25vl,internvl3} designed to progressively build multimodal capability. We construct a high-quality corpus by integrating English-dominant sources with carefully selected Korean additions. The corpus includes multi-image instruction data, culturally contextual content, and safety-critical scenarios (both curated and synthesized). This diverse composition ensures robust cross-modal learning, bilingual competence, and alignment with safety and localization goals. The model is trained on approximately 6.5 billion text tokens and 30.4 billion image tokens. A breakdown of token usage per stage is shown in Table~\ref{tab:training-tokens}, and key training hyperparameters are summarized in Table~\ref{tab:hyperparams}.

\subsubsection{Stage 1. Feature Alignment Pre-training}
To bridge the inherent mismatch between the separately pre-trained vision encoder and language model, we train only the MLP connector to project visual features into the language embedding space. Both the vision encoder and language model remain frozen, while paired image-caption data allow the MLP to learn a reliable alignment between the two modalities. All input images are uniformly resized to a fixed resolution prior to encoding, ensuring consistent visual representations and stable training dynamics.

At this stage, the model is trained to generate textual outputs from images alone, without relying on explicit text prompts. To reduce the impact of noisy image-caption pairs, we use a filtered dataset of real-world images paired with concise, well-formatted English descriptions of their key objects. The structured training dataset enables the model to generate sentences in a consistent output format and learn robust input-output alignment, resulting in strong image-to-text generation capabilities.

\subsubsection{Stage 2. Basic Supervised Fine-tuning}
From this stage onward, we use the AnyRes training strategy introduced in Section~\ref{sec:arch}, which allows flexible handling of input images with varying resolutions. In Stage~2, images are processed at relatively low resolutions to reduce computational overhead, and all model components are jointly trained in single-image settings. To achieve effective instruction tuning across diverse downstream tasks, we focus on building a foundation of broad world knowledge and strong visual-textual understanding.

\begin{itemize}
    \item \textbf{General:} We curate captioning datasets covering real-world images, charts, and tables, with a strong emphasis on quality improvement. In addition to selecting high-value open-source data, we re-caption the datasets in-house using VLMs to enhance accuracy, fluency, and consistency. This re-captioning process helps the model to better acquire knowledge about diverse objects, structural layouts, and reasoning over tabular or graphical information.
    \item \textbf{Text Recognition:} We construct a bilingual text-recognition dataset containing Korean and English text in diverse fonts and styles, composed of both collected samples and synthetically generated images with embedded text. Training prompts are formatted as standardized instructions to enhance robust bilingual text recognition and reading comprehension capabilities (e.g., “OCR this image section by section, from top to bottom, and left to right. Do not insert line breaks in the output text. If a word is split due to a line break in the image, use a space instead.”).
\end{itemize}

\subsubsection{Stage 3. Advanced Supervised Fine-tuning}
In Stage~3, we aim to train the model to handle more complex scenarios and improve spatial precision. In single-image settings, input images are processed at higher resolutions than in the previous phase, producing finer-grained visual representations. For multi-image scenarios, we adopt a fixed-size image representation strategy to keep token lengths manageable and ensure compatibility with the model’s context window. This allows the model to scale to multi-image reasoning tasks without sacrificing visual fidelity. We expand the dataset to support instruction tuning for a wide range of image-based tasks.

\begin{itemize}
    \item \textbf{General:} We construct an image-based QA dataset spanning diverse tasks. To strengthen the model’s bilingual capabilities, we regenerate Korean queries from the original English prompts and their corresponding answers, producing a high-quality Korean image-QA dataset. Furthermore, we employ human annotators to enrich the queries by appending task-specific prompts aligned with the target output format (e.g., “Answer the question using a single word or phrase.”, “Explain the solution process step by step and provide the final answer.”). Exposure to such datasets enhances the model’s instruction-following ability, which in turn increases its usability.
    \item \textbf{Document:} We use an in-house dataset designed for document-based question answering (QA), consisting of up to 12 Korean-English images per sample across a wide range of domains. Since generating QA pairs directly from images can often lead to hallucinations, we adopt two strategies that use text as a reference. The first is to collect the accompanying text when crawling document images and construct QA pairs based on the text. The second strategy is to create new QA pairs from the document text and generate corresponding synthetic images for each document using different templates. This pipeline allows us to minimize hallucination and significantly improve the model’s performance on multi-image document QA tasks.
    \item \textbf{Fine-grained:} Inspired by Kosmos-2~\citep{kosmos2}, we produce grounding and referring datasets. Grounding tasks require the model to identify the locations of objects mentioned in the user query within images. Queries should be annotated with the special \texttt{<gro>} token to prompt the model to perform grounding. In contrast, referring is to provide an appropriate, context-based answer based on the objects designated by the user. The locations of objects are represented in the following format:
\[
\texttt{<obj>\{object\}</obj><bbox>\{x1\}, \{y1\}, \{x2\}, \{y2\}</bbox>}
\]
where \texttt{<obj>} encloses the recognized object text and \texttt{<bbox>} specifies its bounding box as normalized coordinates $(x_{\mathrm{min}}, y_{\mathrm{min}}, x_{\mathrm{max}}, y_{\mathrm{max}})$ within $[0,1]$.

    \item \textbf{OCR:} We further develop fine-grained OCR datasets in which inputs for the model are a single image and the \texttt{<ocr>} query. The model is required to detect and recognize every word in the image. Words are segmented at whitespace boundaries, ensuring consistent unit-level annotation. Each word is annotated using the format:
\[
\texttt{<char>\{word\}</char><bbox>\{x1\}, \{y1\}, \{x2\}, \{y2\}</bbox>}
\]
The recognized text is ordered based on y-coordinate clustering to imitate the human reading pattern, proceeding from top to bottom and left to right. This structured annotation lets the model to demonstrate precise character-level recognition and robust word-level understanding in both Korean and English.
\end{itemize}

\subsubsection{Stage 4. Preference Optimization}
The final stage employs Direct Preference Optimization (DPO)~\citep{dpo}. Unlike the previous model \varcovision-14B, which updates only the LLM layers in Stage~4, \varcovision-2.0 unfreezes the entire model for full end-to-end preference optimization. These improvements lead to more accurate grounding, richer visual understanding, and better alignment with user intent in practical applications. Preference optimization focuses on guiding the model towards safe, accurate, and culturally appropriate responses. We design training datasets for Stage~4 across three categories:

\begin{itemize}
    \item \textbf{General:} Following the approach of MMPR-v1.2~\citep{internvl25mpo}, we develop an in-house dataset specifically for general-purpose images. The goal is to make the model generate precise and reliable answers without hallucination. This dataset serves as the backbone dataset for refining the model’s response quality across a broad spectrum of visual tasks.
    \item \textbf{Safety:} We build datasets targeting safety-critical scenarios, where the model needs to appropriately refuse harmful or unsafe queries, including cases where users submit misleading instructions or provide sensitive images.
    \item \textbf{Localization:} We emphasize cultural and contextual awareness by constructing datasets with Korean-specific images and enriched responses. For instance, an image of Gyeongbokgung Palace is not simply annotated as “a palace,” but as “Gyeongbokgung Palace in Seoul, the main royal palace of the Joseon dynasty.” This enables the model to provide culturally relevant and context-aware answers.
\end{itemize}

\begin{table}[h]
  \caption{Training data usage across stages.}
  \label{tab:training-tokens}
  \centering
  \normalsize
  \widecenter{%
  \begin{tabular}{lccc}
    \toprule
    Stage & Text tokens & Image tokens & Total tokens \\
    \midrule
    1 & 4M   & 560M  & 564M \\
    2 & 760M & 4.7B  & 5.46B \\
    3 & 5.7B & 25B   & 30.7B \\
    4 & 61M  & 93M   & 154M \\
    \midrule
    \textbf{Total} & 6.5B & 30.4B & 36.9B \\
    \bottomrule
  \end{tabular}}
\end{table}

\begin{table}[h]
  \caption{Hyperparameters.}
  \label{tab:hyperparams}
  \centering
  \normalsize
  \begin{tabular}{llcccc}
    \toprule
    Model & Hyperparameter & Stage 1 & Stage 2 & Stage 3 & Stage 4 \\
    \midrule
    \multirow{9}{*}{14B}
      & Trainable     & MLP        & Full Model & Full Model & Full Model \\
      & Batch Size    & 128        & 128        & 128        & 128 \\
      & Context Length& 1024       & 16384      & 16384      & 9216 \\
      & LR            & 1e-3       & 1e-5       & 1e-5       & 3e-7 \\
      & LR (Vision)   & --         & 2e-6       & 2e-6       & 3e-7 \\
      & LR Schedule   & cosine     & cosine     & cosine     & constant \\
      & Max. \#Grids  & 1$\times$1 & 2$\times$2 & 6$\times$6 & 6$\times$6 \\
      & Max. \#Tokens & 576        & (4+1)$\times$576 & (9+1)$\times$576 & (9+1)$\times$576 \\
    \midrule
    \multirow{9}{*}{1.7B}
      & Trainable     & MLP        & Full Model & Full Model & Full Model \\
      & Batch Size    & 128        & 128        & 128        & 128 \\
      & Context Length& 1024       & 16384      & 16384      & 9216 \\
      & LR            & 1e-3       & 1e-5       & 1e-5       & 6e-7 \\
      & LR (Vision)   & --         & 2e-6       & 2e-6       & 6e-7 \\
      & LR Schedule   & cosine     & cosine     & cosine     & constant \\
      & Max. \#Grids  & 1$\times$1 & 2$\times$2 & 6$\times$6 & 6$\times$6 \\
      & Max. \#Tokens & 576        & (4+1)$\times$576 & (9+1)$\times$576 & (9+1)$\times$576 \\
    \bottomrule
  \end{tabular}
\end{table}

\subsection{Initialization Strategy for \varcovision-2.0-1.7B}

The \varcovision-2.0-1.7B model shares the same overall architecture as its 14B counterpart, with the only difference being the use of Qwen3-1.7B as the language model in place of Qwen3-14B. Inspired by Progressive Scaling~\citep{internvl25}, we initialize the vision encoder of the 1.7B model with weights from the 14B model after Stage~3 training. This approach facilitates knowledge transfer from the larger model and accelerates convergence. The impact of this initialization strategy is further discussed in our ablation study (Section~\ref{sec:ablation}).

\subsection{Model Merging}
For the 14B variant, we adopt a merge-train-merge strategy to stabilize training and improve generalization of the model. Inspired by prior work on weight-space model averaging~\citep{swa}, we first merge multiple Stage~3 checkpoints to obtain a robust initializer for Stage~4 (preference optimization). After Stage~4 completes, we perform another round of checkpoint merging to produce the final model. This approach reduces checkpoint variance and aggregates distinct patterns learned across multiple checkpoints without introducing additional inference overhead.

In contrast, we do not apply model merging to the lightweight 1.7B variant, as trials on averaging multiple checkpoints with the 1.7B model have degraded validation performance. We find this is due to the lower effective dimensionality in smaller models, where parameter vectors are less likely to be mutually orthogonal. This property of smaller models makes weight averaging particularly vulnerable to interference. The hypothesis is further supported by recent work that identifies parameter interference as a key limitation in merging low-dimensional models~\citep{ties}. Therefore, we opt for a simpler strategy for the 1.7B model by selecting one single best-performing checkpoint as its final version.

\subsection{Training Infrastructure and Efficiency Optimizations}

We use a single compute node equipped with 8$\times$H100 GPUs. To alleviate memory bottlenecks and enable large-scale training within the constrained setup, we employ several efficiency-oriented strategies.

\begin{itemize}
    \item \textbf{Parallelization and Memory Reduction}: We leverage Fully Sharded Data Parallel (FSDP)~\citep{fsdp} as the primary parallelization strategy. FSDP shards not only model parameters but also optimizer states and gradients across GPUs, enabling substantial memory savings and efficient scaling. To further reduce memory overhead, we use activation checkpointing~\citep{checkpointing} and the 8-bit Adam optimizer~\citep{adam8bit}, which minimize the memory footprint of intermediate activations and optimizer states, respectively.
    \item \textbf{Memory-Efficient Logit Computation:} As modern LLMs adopt extensive vocabulary sizes often exceeding 100K tokens, memory consumption during logit computation has emerged as a major bottleneck in training. \varcovision-2.0 follows this trend with a vocabulary size of approximately 150K, making the logit tensor one of the dominant contributors to peak GPU memory usage. To mitigate this, we integrate the Liger kernel~\citep{liger}, which substantially reduces the memory overhead associated with logit computation.
    \item \textbf{CPU Offloading and Logit Chunking in DPO:} During DPO training, memory demands increase due to the simultaneous use of both the target and reference models. Thus, we apply CPU offloading, shifting portions of model storage to host memory at the cost of added communication latency. Although the Liger kernel proves its effectiveness during SFT, it offers limited benefits in the DPO setting. As a result, we implement a custom logit chunking strategy tailored to DPO, allowing us to manually partition the logit computation and better control peak memory usage.
\end{itemize}

Overall, training \varcovision-2.0 requires approximately 700 hours of wall-clock time on a single 8$\times$H100 node, totaling around 5,600 GPU-hours across all stages.

\section{Experiments}

\subsection{Evaluation}

\subsubsection{English Benchmarks}
We evaluate \varcovision-2.0 against SOTA open-weight VLMs, including InternVL3~\citep{internvl3}, Ovis2~\citep{ovis}, and Qwen2.5-VL~\citep{qwen25vl}, across a broad set of widely adopted English benchmarks. The results are sourced from the OpenCompass VLM leaderboard\footnote{\url{https://rank.opencompass.org.cn/leaderboard-multimodal}, \url{https://huggingface.co/spaces/opencompass/open\_vlm\_leaderboard}} if available, to ensure fair and consistent comparisons. Otherwise, evaluations are conducted using the VLMEvalKit~\citep{vlmevalkit}.

Overall, \varcovision-2.0 demonstrates performance competitive with leading open-weight models (Table~\ref{tab:benchmark-comparison}). It shows strong perceptual capabilities and excels in spatial grounding tasks, particularly in benchmarks emphasizing real-world understanding and low-level visual features (e.g., RealWorldQA~\citep{realworldqa}, SEEDBench\_IMG~\citep{seedbench}, Q-Bench~\citep{qbench}, A-Bench~\citep{abench}). However, performance drops on tasks requiring complex reasoning, scientific knowledge, and document/OCR understanding (e.g., ScienceQA~\citep{scienceqa}, DocVQA~\citep{docvqa}, TextVQA~\citep{textvqa}, OCRBench), suggesting clear directions for future improvement.

As presented in Table~\ref{tab:english-lightweight}, the 1.7B variant delivers the best performance on RealWorldQA among lightweight models, indicating strong spatial perception in real-world contexts. On the other hand, its performance is relatively weaker on knowledge-intensive and document-oriented tasks such as ScienceQA and DocVQA. Although the average score is slightly lower than that of other lightweight models, it displays distinctive strengths in physical-world visual understanding.

\begin{table}[p]
  \caption{English Benchmark Results (Large Models). Since no size-equivalent release of Qwen2.5-VL is available, we report Qwen2.5-VL-7B as a reference. \varcovision~is abbreviated as VV. Best in \textbf{bold}, runner-up \underline{underlined}.}
  \label{tab:benchmark-comparison}
  \centering
  \normalsize
  \widecenter{%
  \begin{tabular}{lccccc}
    \toprule
    Benchmark & InternVL3-14B & Ovis2-16B & Qwen2.5-VL-7B & VV-1.0-14B & VV-2.0-14B \\
    \midrule
    MMStar~\citep{mmstar} & \textbf{68.9} & \underline{67.2} & 64.1 & 64.1 & 66.9 \\
    MMMU\_VAL~\citep{mmmu}& \textbf{64.8} & 60.7 & 58.0 & 56.3 & \underline{61.9} \\
    MathVista~\citep{mathvista}& \textbf{74.4} & \underline{73.7} & 68.1 & 67.6 & 73.2 \\
    OCRBench& 87.7 & \underline{87.9} & \textbf{88.8} & 81.5 & 86.9 \\
    AI2D~\citep{ai2d}& \underline{86.0} & \textbf{86.3} & 84.3 & 83.9 & 85.7 \\
    HallusionBench~\citep{hallusionbench}& \underline{55.9} & \textbf{56.8} & 51.9 & 46.8 & 53.2 \\
    MMVet~\citep{mmvet}& \textbf{80.5} & 68.4 & \underline{69.7} & 53.0 & 68.9 \\
    SEEDBench\_IMG& 77.5 & \underline{77.7} & 77.0 & 76.6 & \textbf{78.0} \\
    LLaVABench~\citep{llava}& 84.4 & \textbf{93.0} & \underline{91.0} & 72.8 & 90.2 \\
    RealWorldQA& 69.8 & \underline{74.1} & 68.4 & 72.5 & \textbf{74.6} \\
    POPE~\citep{pope}& \textbf{89.4} & 87.5 & 85.9 & 87.9 & \underline{89.2} \\
    ScienceQA\_TEST& \textbf{98.6} & 95.2 & 89.0 & \underline{98.5} & 93.5 \\
    SEEDBench2\_Plus~\citep{seedbench2plus}& 70.1 & \textbf{72.1} & 70.7 & 69.9 & \underline{71.9} \\
    BLINK~\citep{blink}& \textbf{59.9} & \underline{59.0} & 55.3 & 50.3 & 54.5 \\
    TextVQA\_VAL& 82.2 & \underline{83.0} & \textbf{85.4} & 59.5 & 80.4 \\
    ChartQA\_TEST~\citep{chartqa}& \textbf{87.8} & 79.1 & 80.6 & 34.2 & \underline{84.2} \\
    Q-Bench1\_VAL& 76.5 & \underline{79.2} & 78.2 & 74.7 & \textbf{79.9} \\
    A-Bench\_VAL& 76.3 & \textbf{79.6} & 75.4 & 74.4 & \underline{79.5} \\
    DocVQA\_TEST& 94.1 & \underline{94.9} & \textbf{95.7} & 77.8 & 90.9 \\
    InfoVQA\_TEST~\citep{infovqa}& \textbf{83.6} & \underline{82.8} & 82.6 & 60.2 & 80.4 \\
    \midrule
    \textbf{Average} & \textbf{78.4} & \underline{77.9} & 76.0 & 68.1 & 77.2 \\
    \bottomrule
  \end{tabular}}
\end{table}

\begin{table}[p]
  \caption{English Benchmark Results (Lightweight Models).}
  \label{tab:english-lightweight}
  \centering
  \normalsize
  \widecenter{%
  \begin{tabular}{lccc}
    \toprule
    Benchmark & InternVL3-2B & Ovis2-2B & VV-2.0-1.7B \\
    \midrule
    MMStar          & \textbf{61.1} & \underline{56.7} & 54.5 \\
    MMMU\_VAL       & \textbf{48.7} & \underline{45.6} & 44.1 \\
    MathVista       & 57.6 & \textbf{64.1} & \underline{61.1} \\
    OCRBench        & \underline{83.1} & \textbf{87.3} & 83.0 \\
    AI2D            & \underline{78.6} & \textbf{82.7} & 76.0 \\
    HallusionBench  & 41.9 & \textbf{50.2} & \underline{43.0} \\
    MMVet           & \textbf{67.0} & \underline{58.3} & 52.7 \\
    SEEDBench\_IMG  & \textbf{75.0} & 74.4 & \underline{74.5} \\
    LLaVABench      & 72.1 & \underline{76.6} & \textbf{77.3} \\
    RealWorldQA     & 65.1 & \underline{66.0} & \textbf{66.8} \\
    POPE            & \textbf{90.1} & 87.8 & \underline{88.6} \\
    ScienceQA\_TEST & \textbf{95.8} & \underline{91.2} & 84.0 \\
    SEEDBench2\_Plus& 64.8 & \textbf{67.4} & \underline{66.9} \\
    BLINK           & \textbf{53.1} & \underline{47.9} & 47.2 \\
    TextVQA\_VAL    & \underline{78.6} & \textbf{80.0} & 77.0 \\
    ChartQA\_TEST   & \underline{76.0} & \textbf{81.4} & 75.7 \\
    Q-Bench1\_VAL   & 71.9 & \textbf{76.3} & \underline{72.3} \\
    A-Bench\_VAL    & \underline{74.3} & \textbf{76.2} & 72.4 \\
    DocVQA\_TEST    & \underline{88.2} & \textbf{91.9} & 83.5 \\
    InfoVQA\_TEST   & \underline{66.9} & \textbf{71.7} & 65.0 \\
    \midrule
    \textbf{Average} & \underline{70.5} & \textbf{71.7} & 68.3 \\
    \bottomrule
  \end{tabular}}
\end{table}

\subsubsection{Korean Benchmarks}

To evaluate Korean language capabilities, we assess \varcovision-2.0 on a set of publicly available Korean multimodal benchmarks: K-MMBench, K-MMStar, K-SEED, K-LLaVA-W, and K-DTCBench~\citep{varcovision}. The same baseline models used in the English evaluation are included for consistency. All evaluations are conducted using the original Korean instructions provided in each benchmark.

In Table~\ref{tab:korean-benchmark}, \varcovision-2.0 demonstrates the highest overall average among all compared models, proving its strong generalization across a wide range of Korean tasks. In particular, it shows substantial gains on K-LLaVA-W, indicating superior capabilities in Korean text generation and dialogue understanding. On the other hand, performance on K-DTCBench remains relatively weaker, likely due to the benchmark's emphasis on structured data understanding involving documents, tables, and charts. This aligns with the domain-specific limitations in the English benchmarks.

We find that Ovis2-16B shows unusually low scores on K-MMStar, which appear to result from the model frequently failing to follow the expected response format. This suggests that the performance drop arises from challenges with Korean instruction formats rather than from limitations in language understanding. To validate the hypothesis, we re-evaluate all models using the English instruction provided by VLMEvalKit. The results are discussed in the Appendix~\ref{app:kor-bench-eng-inst}.

As reported in Table~\ref{tab:korean-lightweight}, the 1.7B variant ranks highest among the lightweight models. It performs strongly on K-MMBench\_DEV, K-SEED, and K-LLaVA-W, showcasing solid comprehension and generation capabilities in Korean. Nonetheless, its performance on K-DTCBench lags behind, reflecting challenges in structured visual reasoning once again.

The 1.7B model also underperforms on K-MMStar, primarily owing to frequent response-format violations under Korean instructions (e.g., generating free-form answers instead of selecting the correct choice letter). Notably, this issue does not occur with the 14B model. We propose two possible factors contributing to the 1.7B model’s low performance: (i) the absence of Korean multiple-choice data in the Stage~3 training corpus, which limits exposure to the task schema, and (ii) the greater capacity of the 14B model, which enables better generalization to unseen instruction formats, reducing such errors despite equivalent training data setup.

\begin{table}[h]
  \caption{Korean Benchmark Results (Large Models).}
  \label{tab:korean-benchmark}
  \centering
  \normalsize
  \widecenter{%
  \begin{tabular}{lccccc}
    \toprule
    Benchmark & InternVL3-14B & Ovis2-16B & Qwen2.5-VL-7B & VV-1.0-14B & VV-2.0-14B \\
    \midrule
    K-MMBench\_DEV & \textbf{89.1} & 86.0 & 84.7 & 84.8 & \underline{87.7} \\
    K-MMStar       & \textbf{64.9} & 29.7 & 49.3 & 58.8 & \underline{63.6} \\
    K-SEED         & \textbf{78.2} & 73.2 & 75.7 & 75.4 & \underline{77.2} \\
    K-LLaVA-W      & 80.9 & 86.3 & \underline{94.1} & 83.1 & \textbf{96.5} \\
    K-DTCBench     & \textbf{87.9} & 81.7 & 82.1 & \underline{84.6} & 78.3 \\
    \midrule
    \textbf{Average} & \underline{80.2} & 71.4 & 77.2 & 77.3 & \textbf{80.7} \\
    \bottomrule
  \end{tabular}}
\end{table}

\begin{table}[h]
  \caption{Korean Benchmark Results (Lightweight Models).}
  \label{tab:korean-lightweight}
  \centering
  \normalsize
  \widecenter{%
  \begin{tabular}{lccc}
    \toprule
    Benchmark & InternVL3-2B & Ovis2-2B & VV-2.0-1.7B \\
    \midrule
    K-MMBench\_DEV & \underline{76.9} & 68.4 & \textbf{77.9} \\
    K-MMStar       & \textbf{50.1}    & 10.9 & \underline{40.8} \\
    K-SEED         & \underline{69.2} & 34.5 & \textbf{70.7} \\
    K-LLaVA-W      & 47.6             & \underline{67.2} & \textbf{73.5} \\
    K-DTCBench     & \textbf{68.8}    & 44.6 & \underline{64.2} \\
    \midrule
    \textbf{Average} & \underline{62.5} & 45.1 & \textbf{65.4} \\
    \bottomrule
  \end{tabular}}
\end{table}

\subsubsection{Text-only Benchmarks}

We view VLMs as language models extended with visual understanding. From this perspective, our primary goal is to preserve the model’s linguistic foundation and demonstrate that \varcovision\ maintains strong performance on text-only benchmarks. As shown in Tables~\ref{tab:textonly-large} and~\ref{tab:textonly-light}, both the full-scale and lightweight variants of \varcovision-2.0 achieve consistently high scores across various text-only tasks, including general knowledge, instruction following, and multi-turn reasoning. These results indicate that the language capabilities have been effectively maintained throughout our training phases.

We credit this performance largely to the strength of the underlying language foundation model, Qwen3. Furthermore, unlike other models that depend on massive amounts of multimodal data, \varcovision-2.0 relies on more compact, high-quality training datasets and efficient learning strategies. This implies that strong performance does not necessarily require maximal training data. Instead, fine-grained data curation and optimized training strategies may be more critical for enhancing text-only capabilities.

We observe that some models show anomalously low performance on the MMLU benchmark~\citep{mmlu} because of the response-format violations, as also seen in other multimodal benchmarks. These cases do not indicate significant deficiencies in the models’ knowledge or reasoning capabilities. We encourage readers to keep this in mind when interpreting benchmark results.

\begin{table}[h]
  \caption{Text-only Benchmark Results (Large Models).}
  \label{tab:textonly-large}
  \centering
  \normalsize
  \widecenter{%
  \begin{tabular}{lccccc}
    \toprule
    Benchmark & InternVL3-14B & Ovis2-16B & Qwen2.5-VL-7B & VV-1.0-14B & VV-2.0-14B \\
    \midrule
    MMLU       & \textbf{78.5} & \underline{78.4} & 4.6   & 4.9 & 77.9 \\
    MT-Bench~\citep{mtbench}& \underline{89.3} & 85.9 & 80.7 & 87.7 & \textbf{89.8} \\
    KMMLU~\citep{kmmlu}& \underline{51.4} & 49.3 & 39.6 & 37.7 & \textbf{57.5} \\
    KoMT-Bench~\citep{komtbench}& 70.1 & \underline{79.1} & 68.4 & \textbf{83.8} & 78.3 \\
    LogicKor~\citep{logickor}& 70.0 & \underline{79.4} & 65.5 & \textbf{86.7} & 74.0 \\
    \midrule
    \textbf{Average} & 71.9 & \underline{74.4} & 51.7 & 60.2 & \textbf{75.5} \\
    \bottomrule
  \end{tabular}}
\end{table}

\begin{table}[h]
  \caption{Text-only Benchmark Results (Lightweight Models).}
  \label{tab:textonly-light}
  \centering
  \normalsize
  \widecenter{%
  \begin{tabular}{lccc}
    \toprule
    Benchmark & InternVL3-2B & Ovis2-2B & VV-2.0-1.7B \\
    \midrule
    MMLU       & \textbf{59.9} & 12.9 & \underline{55.3} \\
    MT-Bench   & \underline{62.8} & 61.4 & \textbf{72.3} \\
    KMMLU      & \textbf{38.0} & \underline{31.1} & 10.4 \\
    KoMT-Bench & 29.1 & \underline{34.4} & \textbf{59.1} \\
    LogicKor   & 25.6 & \underline{31.2} & \textbf{53.7} \\
    \midrule
    \textbf{Average} & \underline{43.1} & 34.2 & \textbf{50.2} \\
    \bottomrule
  \end{tabular}}
\end{table}

\begin{table}[!h]
  \caption{Korean Cultural Benchmark Results (Large Models).}
  \label{tab:cultural-large}
  \centering
  \normalsize
  \widecenter{%
  \begin{tabular}{lccccc}
    \toprule
    Benchmark & InternVL3-14B & Ovis2-16B & Qwen2.5-VL-7B & VV-1.0-14B & VV-2.0-14B \\
    \midrule
    K-Viscuit~\citep{kviscuit}& 71.7 & \textbf{77.0} & 70.9 & 69.3 & \underline{73.7} \\
    PangeaBench (ko)~\citep{pangea} & \textbf{77.2} & \underline{76.9} & 76.6 & 67.6 & 74.5 \\
    \midrule
    \textbf{Average} & \underline{74.5} & \textbf{77.0} & 73.8 & 68.5 & 74.1 \\
    \bottomrule
  \end{tabular}}
\end{table}

\begin{table}[!h]
  \caption{Korean Cultural Benchmark Results (Lightweight Models).}
  \label{tab:cultural-light}
  \centering
  \normalsize
  \widecenter{%
  \begin{tabular}{lccc}
    \toprule
    Benchmark & InternVL3-2B & Ovis2-2B & VV-2.0-1.7B \\
    \midrule
    K-Viscuit        & \underline{60.0} & \textbf{64.1} & 57.7 \\
    PangeaBench (ko) & \textbf{66.2} & 63.1 & \underline{63.8} \\
    \midrule
    \textbf{Average} & \underline{63.1} & \textbf{63.6} & 60.8 \\
    \bottomrule
  \end{tabular}}
\end{table}

\subsubsection{Korean Cultural Benchmarks}

We also evaluate \varcovision-2.0 on benchmarks assessing understanding of Korean culture and region-specific contexts (Tables~\ref{tab:cultural-large} and~\ref{tab:cultural-light}). While the model shows improvements over its previous version, its performance remains less competitive compared to other leading models. We attribute this performance gap to the relatively limited availability of training data related to Korean culture, which constrains the model’s capacity to understand cultural contexts. Expanding and diversifying this portion of the training corpus presents a promising direction for future work.

\subsubsection{OCR Benchmarks}
\label{sec:ocr}

We evaluate \varcovision-2.0's OCR capabilities on three datasets: CORD~\citep{cord}, ICDAR2013~\citep{icdar2013}, and ICDAR2015~\citep{icdar2015}. These benchmarks require not only accurate recognition of textual content but also precise spatial localization within images.

We find that upscaling input images to a minimum resolution of 2,304 pixels on the longer side (for smaller images) significantly improves the OCR performance. The SigLIP2 vision encoder~\citep{siglip2} used in our model operates with an input resolution of 384, and training is performed with a maximum spatial grid size of $6 \times 6$. This implies that the effective maximum resolution is $384 \times 6 = 2,304$, which corresponds to the most fine-grained tokenization supported by the model. Maximizing the number of visual tokens enables the model to capture visual signals in greater detail, and we therefore adopt this resolution for all reported results.

As presented in Table~\ref{tab:ocr}, \varcovision-2.0 shows substantial gains over popular open-source OCR systems such as PaddleOCR~\citep{paddleocr} and EasyOCR~\citep{easyocr}, achieving notably higher accuracy across all benchmarks. When compared to CLOVA OCR~\citep{naverCLOVAocr}—a strong commercial OCR system—our model demonstrates competitive performance, outperforming it on CORD and ICDAR2013, and closely approaching its accuracy on ICDAR2015. These results are particularly noteworthy given that \varcovision-2.0 is not explicitly trained as an OCR-specific model, which highlights its strong visual-textual alignment and potential for broader real-world applications beyond conventional OCR.

\begin{table}[t]
  \caption{OCR benchmark Results (Recognition Accuracy).}
  \label{tab:ocr}
  \centering
  \normalsize
  \widecenter{%
  \begin{tabular}{lcccccc}
    \toprule
    Benchmark & CLOVA OCR & PaddleOCR & EasyOCR & VV-1.0-14B & VV-2.0-1.7B & VV-2.0-14B \\
    \midrule
    CORD      & 93.9 & 91.4 & 77.8 & 81.9 & \underline{96.2} & \textbf{97.1} \\
    ICDAR2013 & 94.4 & 92.0 & 85.0 & 94.4 & \textbf{95.9} & \underline{95.7} \\
    ICDAR2015 & \textbf{84.1} & 73.7 & 57.9 & 73.5 & 73.7 & \underline{79.4} \\
    \midrule
    \textbf{Average} & \textbf{90.8} & 85.7 & 73.6 & 83.3 & 88.6 & \underline{90.7} \\
    \bottomrule
  \end{tabular}}
\end{table}

\subsection{Ablation Study}
\label{sec:ablation}

Due to compute budget constraints, we only conduct ablations on the 1.7B variant using the eight main benchmarks from the OpenCompass VLM leaderboard. Unless otherwise specified, we follow the same training recipe as in the main setting and report average performance across benchmarks.

\subsubsection{Vision Encoder Sharing}
As described earlier, the 1.7B model initializes its vision encoder with weights from the 14B model trained up to Stage3. Compared to off-the-shelf SigLIP2 initialization, this shared-encoder initialization results in higher average performance (Table\ref{tab:ablation-sharing}). We also experiment with freezing the vision encoder and training only the remaining components of the 1.7B model to evaluate whether this preserves the encoder’s original capabilities throughout the four-step training process. However, this approach leads to lower performance than full end-to-end training, highlighting the importance of continued joint optimization even when transferring from a strong encoder.

\begin{table}[h]
\vspace{2mm}
  \caption{Ablation on the 1.7B model: vision-encoder sharing/freezing. 
  Abbrev.: NS = not shared; SH = shared; FRZ$\le$2/3 = freeze vision-encoder until Stage~2/3.}
  \label{tab:ablation-sharing}
  \centering
  \normalsize
   \resizebox{\textwidth}{!}{
  \begin{tabular}{lccccccccc}
    \toprule
    Setting & MMBv1.1 & MMStar & AI2D & MMMU & MathVista & HallusionBench & OCRBench & MMVet & Avg. \\
    \midrule
    NS    & 73.9 & 51.7 & 74.6 & 43.0 & 56.6 & 35.3 & 79.4 & 50.4 & 58.1 \\
    SH    & 74.5 & 54.6 & 75.8 & 44.2 & 61.8 & 41.7 & 82.2 & 49.3 & \textbf{60.5} \\
    FRZ$\le$2 & 72.8 & 52.4 & 75.2 & 42.6 & 55.1 & 36.9 & 78.6 & 51.9 & 58.2 \\
    FRZ$\le$3 & 72.7 & 50.7 & 73.8 & 42.7 & 57.4 & 36.4 & 76.5 & 50.8 & 57.6 \\
    \bottomrule
  \end{tabular}}
\end{table}

\subsubsection{DPO Variant Experiments}
Inspired by MPO~\citep{internvl25mpo}, we also investigate a combined objective (DPO+SFT) as an alternative to standard DPO-only for preference optimization (Table~\ref{tab:ablation-dpo}). (Note: BCO is excluded from this round of experiments due to engineering constraints, such as logit chunking.) Across the eight main OpenCompass benchmarks, the combined objective setting does not outperform DPO-only setting. In fact, DPO-only model achieves marginally better average performance, though the difference may not be significant. A plausible explanation is that our preference dataset is not well aligned with the SFT-style objective, and the auxiliary loss term may introduce gradient interference that weakens the preference learning signal. Based on these findings, our recipe for preference optimization employs shared-encoder initialization, end-to-end training, and DPO objective.

\begin{table}[h]
\vspace{2mm}
  \caption{Ablation on the 1.7B model: DPO variants.}
  \label{tab:ablation-dpo}
  \centering
  \normalsize
   \resizebox{\textwidth}{!}{
  \begin{tabular}{lccccccccc}
    \toprule
    Setting & MMBv1.1 & MMStar & AI2D & MMMU & MathVista & HallusionBench & OCRBench & MMVet & Avg. \\
    \midrule
    SFT & 74.6 & 53.9 & 75.7 & 42.6 & 61.4 & 43.7 & 81.0 & 50.3 & 60.4 \\
    DPO & 75.0 & 54.5 & 76.0 & 44.1 & 61.1 & 43.0 & 83.0 & 52.7 & \textbf{61.2} \\
    DPO+SFT & 75.0 & 55.0 & 75.7 & 44.6 & 61.0 & 43.3 & 83.1 & 48.6 & 60.8 \\
    \bottomrule
  \end{tabular}}
\end{table}

\subsection{Other Experiments}

\textbf{Extrapolation to Larger Grids and Token Lengths.}
As we find that the number of visual tokens and grid size affect the model's OCR performance in Section \ref{sec:ocr}, we further explore whether the model could extrapolate beyond its original training configuration. When increasing the maximum grid size to $8 \times 8$ and the token sequence length to $(16{+}1) \times 576$, the model exhibits improved performance on OCRBench~\citep{ocrbench} compared to the original setting. However, for OCR tasks requiring precise text localization, performance drops sharply, with accuracy occasionally reaching zero. This suggests that while content-level understanding extrapolates, spatial generalization remains limited.

\textbf{Persistence of Qwen3 Thinking Mode.}
Qwen3~\citep{qwen3} offers a special “thinking mode”, activated by the `\texttt{/think}' flag appended to user queries. We examine if this language model's inherent thinking capability persists even after VLM training phases. However, our experiments with \varcovision-2.0 show no evidence that the thinking mode behavior has transferred to our model.

\textbf{Prompted Reasoning Behavior.}
Although our training data includes only a small set of reasoning-style supervision tasks, we evaluate whether the model might benefit from step-by-step prompting strategies. Specifically, we test prompts such as “Let’s think step by step”~\citep{thinkstepbystep}. We find that these reasoning cues do not lead to improved accuracy. Even when explicitly instructed to follow multi-step reasoning, the model often remains confident in its initial predictions and rarely changes its outputs. We provide several examples of the explicit reasoning instructions we tested in Appendix~\ref{app:reasoning-instructions}.

\section{Limitations}

\textbf{Instruction Robustness.}
Outputs are sensitive to superficial formatting changes (e.g., whitespace, newlines). This suggests an over-reliance on fixed prompt templates and insufficient exposure to diverse instruction formats during training.

\textbf{Knowledge and Document Understanding.}
Despite strong performance in perception and spatial reasoning, the model underperforms on knowledge-intensive and document-centric tasks. We attribute this to limited inclusion of curated knowledge sources and a lack of robust layout-aware supervision during training.

\textbf{Weakened Referring Capability.}
The model shows reduced performance on referring tasks compared to the previous \varcovision~model. Referring tasks require identifying and resolving references to specific objects or regions within images, and it is essential for embodied agents and multimodal systems that must interact with environments based on visual understanding. The current weakness may limit the model’s performance in practical applications involving precise object selection or instruction following grounded in visual context.

\section{Future Work}

\textbf{Small-Model Distillation.}
Prior studies indicate that small models often benefit more from knowledge distillation than from direct supervised fine-tuning~\citep{distil, distilbert, distilvlm}. Based on this, we plan to explore teacher-student training regimes, using the larger model as a teacher for the smaller variant.

\textbf{Reasoning Improvements.}
We look forward to enhancing multi-step reasoning capabilities by incentivizing reasoning behavior through reinforcement learning, as explored in recent studies such as DeepSeek-R1~\citep{deepseekr1}. This involves techniques like chain-of-thought distillation~\citep{llmselfimprove, teachslm} and preference-based fine-tuning~\citep{rlhf}.

\textbf{Efficient Context Handling for High-Resolution and Long-Horizon Inputs.}
Extending the model’s context length is necessary to support high-resolution multi-image inputs and long-horizon video understanding. However, obtaining sufficient training data for very long contexts is challenging, and even when available, learning with such extended sequences greatly increases memory costs. To address this, we plan to adopt techniques such as YaRN~\citep{yarn} for efficient context extension and leverage sequence parallelism~\citep{contextparallel} to reduce memory overhead. In addition, we aim to reduce the number of visual tokens—currently about five times larger than that of text—by exploring strategies such as pixel unshuffling~\citep{shuffle}, which can improve overall context efficiency without sacrificing visual fidelity.

\textbf{Extending to Video Modality and Beyond.}
A key goal is to enhance the model’s capacity for long-horizon video understanding through spatiotemporal encoders, including support for 3D free-viewpoint video~\citep{fvv} with controllable camera trajectories. Additionally, we aim to develop omni-modal models that incorporate modalities such as audio and speech, allowing the model to interpret diverse and complex signals.

\textbf{Toward Embodied Multimodal Agents.}
The AI community is increasingly developing agents that both perceive and act, including interacting with GUIs and manipulating on-screen environments (e.g., Anthropic’s Claude 3.5 Sonnet ‘computer use’ feature~\citep{anthropicComputerUse}). Recent work, such as UI-TARS, shows that end-to-end GUI agents can integrate perception and action within a single model~\citep{uitars}, highlighting a shift toward embodied agents that tightly couple vision, language, and action across digital and physical environments. With \varcovision~models, we hope to gradually contribute to the development of embodied agents. 

\textbf{Scaling.}
We plan to scale both model capacity and data volume to enhance generalization and robustness. This includes adopting efficient architectural designs, expanding the multimodal training corpus with higher-quality and more diverse supervision, and refining preference learning through improved reward signals and scalable algorithms. Together, these efforts would yield more capable, aligned, and deployment-ready multimodal models.

\section{Conclusion}
In this technical report, we present \varcovision-2.0, a Korean-specialized, open-weight VLM available in two sizes (14B and 1.7B). Built on LLaVA-OneVision with a four-stage training curriculum, the models achieve competitive results across multiple tasks compared to leading open-weight VLMs. Their strengths include real-world perception, multi-image understanding, and high-fidelity OCR with bounding boxes, making them well-suited for practical applications in both Korean and English contexts. Text-only benchmark results further demonstrate that the \varcovision~series retains core language capabilities, preserving its foundation as a language model. While the 14B model ranks 8th on the VLM leaderboard among models of comparable size, the lightweight 1.7B model provides a practical option for on-device deployment. Overall, \varcovision-2.0 offers a competitive, efficient, and culturally aware foundation for building practical multimodal systems.

\newpage
{
\small

\bibliographystyle{unsrtnat}
\bibliography{vv2}

}


\appendix

\newpage
\section{Korean Benchmarks under English Instructions}
\label{app:kor-bench-eng-inst}

In this additional experiment, we re-run the evaluations in Table~\ref{tab:korean-benchmark} using English instructions instead of the original Korean ones. This setting is designed to exclude the effects of Korean-specific instructions and directly assess models’ underlying Korean understanding ability. As shown in the results, Ovis2-16B performs even better than InternVL3-14B under this setup (Table~\ref{tab:kor-bench-eng-inst}), in contrast to the outcomes with Korean instructions. This observation suggests that benchmark scores obtained under a single instruction style may not fully represent a model’s overall language understanding ability. It also highlights that many open-weight models still lack robustness to diverse instruction formats.

\begin{table}[h]
  \caption{Korean benchmarks (large models, English instruction setting).}
  \label{tab:kor-bench-eng-inst}
  \centering
  \normalsize
  \widecenter{%
  \begin{tabular}{lcccc}
    \toprule
    Benchmark & InternVL3-14B & Ovis2-16B & Qwen2.5-VL-7B & VV-2.0-14B \\
    \midrule
    K-MMBench\_DEV & \textbf{83.2} & \textbf{83.2} & 78.7 & 82.4 \\
    K-MMStar       & \textbf{66.2} & 64.3 & 61.7 & \underline{65.1} \\
    K-SEED         & 77.7 & \textbf{78.4} & 76.0 & \underline{78.2} \\
    K-LLaVA-W   & 70.2 & \underline{80.9} & 79.8 & \textbf{93.5} \\
    K-DTCBench     & \underline{88.8} & 85.4 & \textbf{90.4} & 81.3 \\
    \midrule
    \textbf{Average} & 75.8 & \underline{76.8} & 75.7 & \textbf{79.2} \\
    \bottomrule
  \end{tabular}}
\end{table}

\section{Explicit Reasoning Instructions}
\label{app:reasoning-instructions}

We test several explicit reasoning instructions to examine whether prompting could encourage the model to refine its answers. Below we list three representative examples.

\lstset{
  basicstyle=\ttfamily\small,
  breaklines=true,
  frame=single,
}
\begin{lstlisting}[caption={Reasoning Prompt 1}]
First, provide a direct answer to the question based on your first impression or intuition.
Then, describe the image in relation to the question.
Provide a more informed answer based on your description.
If the answer may be flawed or could be improved, critique it.
Finally, revise and present the final version of your answer if necessary.

Use the following format. Include only the parts that are relevant or necessary:

Question: 
Direct Answer: 
Description: 
Description-based Answer: xxx
Critique: (if applicable)
Final Answer: (if applicable)

Question: 
\end{lstlisting}

\begin{lstlisting}[caption={Reasoning Prompt 2}]
Step #1: Suggest at least two answer candidates. Think out loud.
Candidates can be single words, short phrases, or full sentences
Step #2: Explain which candidate makes the most sense overall.
Step #3: Restate the question in your own words while keeping its original intent and think carefully about its core meaning before answering.
Step #4: Provide your final answer in detail - give your best guess even if you're unsure.

Use the following format:

Answer Candidates:
- Thinking process for candidate 1: xxx
  Candidate 1: xxx
- Thinking process for candidate 2: xxx
  Candidate 2: xxx
...
Comprehensive Analysis: xxx
Restated Question: xxx
Final Answer: xxx

Question:
\end{lstlisting}

\begin{lstlisting}[caption={Reasoning Prompt 3}]
Step #1: Provide a quick guess.
Step #2: Write down your thinking as you work toward the answer.
Step #3: Provide your answer based on your thought.
Step #4: Review and adjust your answer if needed.
Step #5: Provide your final answer in free form.

Use the following format:

Quick Guess: ...
Thinking process:
...
Answer: ...
Review:
...
Free form Answer:
...

Question:
\end{lstlisting}


\end{document}